\documentclass[letterpaper, 10 pt, conference]{ieeeconf}  
\IEEEoverridecommandlockouts                              
\usepackage{amsmath} 
\usepackage{bm}
\usepackage{graphicx}
\usepackage{tabularx}
\usepackage{url}
\usepackage{xcolor}
\usepackage{booktabs}

\usepackage[style=ieee, maxnames=1, minnames=1, backend=biber, doi=false, url=false, isbn=false, eprint=false]{biblatex}
\title{\LARGE \bf
Optimized Area Coverage in Disaster Response Utilizing Autonomous UAV Swarm Formations 
}

\author{Lampis Papakostas$^{1}$,
        Aristeidis Geladaris$^{1,2}$, 
        Athanasios Mastrogeorgiou$^{1,3}$,~\IEEEmembership{IEEE Student Member},\\
        Jim Sharples$^{4}$,
        Gautier Hattenberger$^{4}$,
        Panagiotis Chatzakos$^{1,5}$,
        Panagiotis Polygerinos$^{*,2}$,~\IEEEmembership{IEEE Senior Member}
\thanks{$^{1}$Lampis Papakostas, Aristeidis Geladaris, Athanasios Mastrogeorgiou and Panagiotis Chatzakos are with Tech Hive Labs, Athens, Greece. (e-mail: \{lampis.papakostas, aris.geladaris, athanasios.mastrogeorgiou, panagiotis.chatzakos\}@thlabs.eu)} %
\thanks{$^{2}$Aristeidis Geladaris and Panagiotis Polygerinos are with the Control Systems and Robotics Lab (CSRL), Mechanical Engineering Dept., School of Engineering, Hellenic Mediterranean University, Heraklion, Crete, Greece. (email: polygerinos@hmu.gr)}%
\thanks{$^{3}$Athanasios Mastrogeorgiou is with the Control Systems Lab (CSL), School of Mechanical Engineering,
National Technical University of Athens, Greece. (email: amast@central.ntua.gr)}%
\thanks{$^{4}$Jim Sharples and Gautier Hattenberger are with the Fédération ENAC ISAE-SUPAERO ONERA, Université de Toulouse, Toulouse, France. (email: \{jim.sharples, gautier.hattenberger\}@enac.fr)}%
\thanks{$^{5}$Panagiotis Chatzakos is with the AI Innovation Centre, University of Essex, Wivenhoe Park, Colchester CO4 3SQ, UK. (email: p.chatzakos@essex.ai)}%
\thanks{*Corresponding author.}
}

\begin{document}

\maketitle
\thispagestyle{empty}
\pagestyle{empty}

\begin{abstract}
This paper presents a UAV swarm system designed to assist first responders in disaster scenarios like wildfires. 
By distributing sensors across multiple agents, the system extends flight duration and enhances data availability, reducing the risk of mission failure due to collisions. 
To mitigate this risk further, we introduce an autonomous navigation framework that utilizes a local Euclidean Signed Distance Field (ESDF) map for obstacle avoidance while maintaining swarm formation with minimal path deviation. 
Additionally, we incorporate a Traveling Salesman Problem (TSP) variant to optimize area coverage, prioritizing Points of Interest (POIs) based on preassigned values derived from environmental behavior and critical infrastructure. 
The proposed system is validated through simulations with varying swarm sizes, demonstrating its ability to maximize coverage while ensuring collision avoidance between UAVs and obstacles.
\end{abstract}

\section{INTRODUCTION}
Uncontrolled fires in Europe cause severe environmental and economic damage, especially in the Mediterranean. Despite decades of data collection, the lack of harmonized information has impeded a unified approach to forest fire prevention \cite{EFFIS}.
Unmanned Aerial Vehicles (UAVs) have emerged as valuable tools for disaster monitoring and management \cite{MohdDaud2021ApplicationsOD}. 
They are instrumental in disaster scenarios requiring rapid damage assessment, like fire detection and monitoring in forest fires \cite{Restas2015DroneAF}, \cite{Glantz2020UAVUI}, \cite{Naser2020ConceptsAA} especially where traditional methods fail due to high-resolution demands, hazardous environments, or time constraints \cite{Erdelj2016Applications}, \cite{Hildmann2019ReviewUU}. 

\begin{figure}[t]
    \centering
    \includegraphics[width=0.48\textwidth]{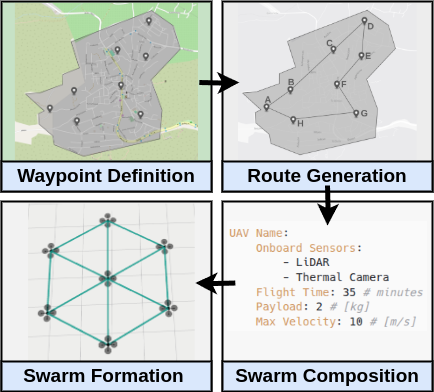}
    \caption{System Overview. User-defined waypoints are prioritized, and a global trajectory is generated using a TSP variant. The UAV swarm is configured based on agent capabilities to maintain an optimal formation.}
    \label{fig:system_design.png}
\end{figure}

However, UAVs face significant limitations, including restricted flight duration and operational range due to energy constraints \cite{app132111932}, limited payload capacity, and potential communication failures caused by signal interference and range limitations \cite{tegicho}. Battery constraints necessitate trade-offs between sensor usage and flight time, while the risk of collisions threatens equipment integrity and mission success.

UAV swarms enhance efficiency by distributing sensors across multiple agents, extending flight time, improving data acquisition, and increasing system resilience \cite{campion}. 
Multiple lower-cost sensors often outperform a single ultra-high-definition (UHD) sensor in field-of-view, cost, and fault tolerance. While UHD sensors offer high detail, they require more power and have limited coverage. In contrast, multiple sensors cover wider areas, reduce movement needs, and enable real-time data fusion. From an economic aspect, they provide redundancy and lower energy consumption, making them ideal for disaster monitoring, industrial inspections, and autonomous navigation \cite{Zhu2024HighPH}. 
UAV swarms also aid first responders by establishing communication networks \cite{Chandran2024MultiUAVNF}, \cite{Erdelj2017Helpfromthesky} and collecting real-time fire data \cite{Liu2023DOMEDM}, enhancing wildfire monitoring and firefighting efficiency \cite{Rao2021DeploymentOW}, \cite{Afgah2019WildfireMI}.

In disaster-affected areas, swarms of autonomous UAVs predominantly follow a leader-follower approach, making them vulnerable to single-point failures. To the best of our knowledge, existing decentralized UAV swarm systems capable of maintaining formation while avoiding obstacles have not yet been applied to disaster management. In \cite{Saffre2022MonitoringAC}, the impact of swarm size, dynamics, and area coverage on wildfire containment is examined, though the system requires human supervision and struggles in cluttered environments. Likewise, \cite{Simoes2020ForestFM} presents a sensor-equipped UAV system that positions drones based on disaster management team locations for fire monitoring. While it incorporates an anti-collision mechanism for UAV-to-UAV avoidance, it lacks comprehensive obstacle-avoidance capabilities.
All in all, key challenges persist in collision avoidance, swarm coordination, networking, security, and limitations related to size, weight, and power \cite{Shakhatreh2018UnmannedAV}, \cite{Khan2022Emerging}. Efforts to optimize swarm algorithms for area coverage and path planning have also gained attention \cite{Sundarrajan2023TheSC}, \cite{Akshya2020GraphbasedPP}, \cite{BandeiraAnalysisOP}.

To address these limitations, we propose a swarm navigation framework that enables coordinated UAV movement between Points of Interest (POIs), ensuring efficient coverage and real-time collision avoidance through inter-UAV communication. A spatial formation strategy is also implemented to maintain systematic coverage and facilitate smoother task execution, particularly for disaster management applications. By distributing power and sensor responsibilities, the
system extends operational range, enhances data collection efficiency, and ensures robustness in dynamic and
challenging environments. Finally, a variant of the Traveling Salesman Problem (TSP) is employed to optimize UAV routing by minimizing travel costs while incorporating a priority-based scoring system for POIs, ensuring an optimal balance between efficiency and mission-critical priorities.

We summarize our key contributions as:
\begin{itemize}
    \item \textit{Development of a swarm navigation schema}: Enables coordinated UAV movement between POIs while ensuring efficient area coverage and robust collision avoidance through inter-UAV communication.
    \item \textit{Implementation of a priority-based TSP variant}: Minimizes travel costs by optimizing routes and incorporates constraints such as time windows and POI importance.
\end{itemize}

The work is organized as follows: Section II outlines the disaster scenario and details the UAV swarm system design. Section III covers the mapping and path-planning algorithms that navigate the swarm formation. Section IV presents simulation results. Finally, Section V offers conclusions and discussions for future directions.


\section{DISASTER ANALYSIS AND SYSTEM DESIGN}
Defining the design parameters and the key components of the proposed system requires an understanding of disaster mechanisms and first responders' strategies. Emergency response operations vary based on disaster type and local service structures. Firefighters, police, and medical teams operate in overlapping zones, with fluctuating perimeters dictated by fire spread, smoke, and other hazards. Enhanced situational awareness is critical during the initial "chaos phase," enabling a targeted response. Real-time, high-resolution disaster overview is essential for assessing damage, prioritizing POIs, and establishing safe zones. In fire incidents, this typically follows a sequence: evaluating disaster extent, identifying triage spots, defining safe zones, and incorporating additional sensor data such as gas detection.
\begin{figure}[t]
    \centering
    \includegraphics[width=0.48\textwidth]{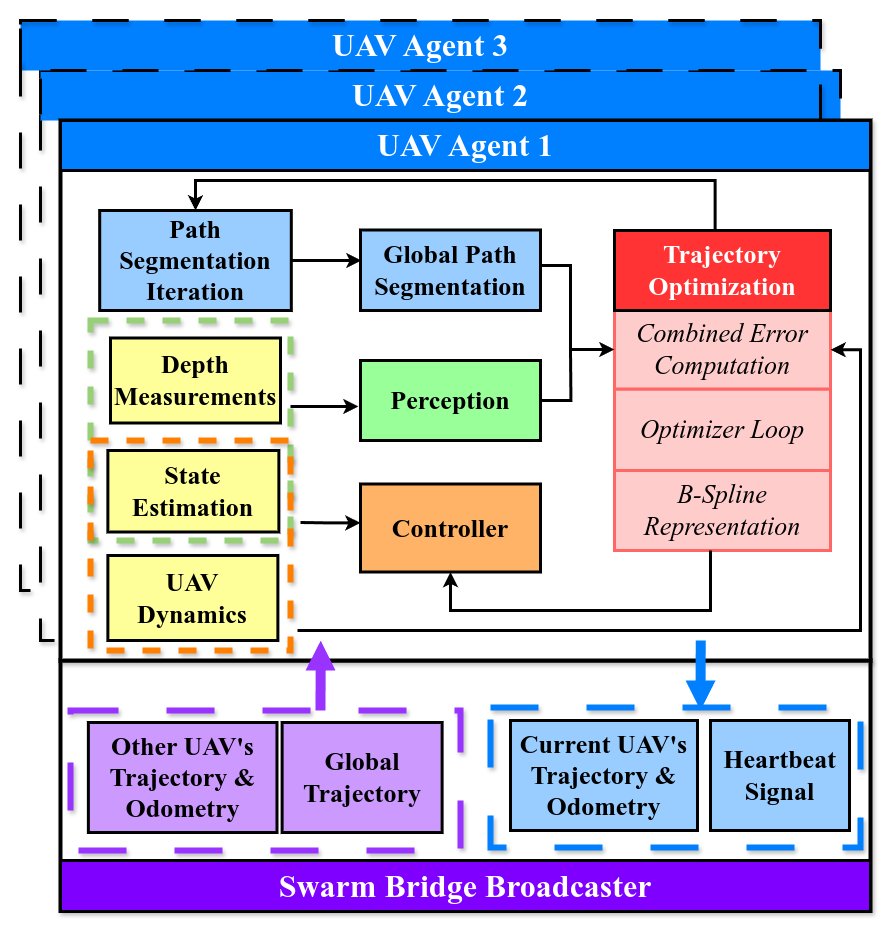}
    \caption{Software architecture of UAV agent. The information exchange between agents includes the modules of localization, perception, control, and trajectory generation.}
    \label{fig:system_architecture.png}
\end{figure}

\begin{figure*}[t]
    \centering
    \includegraphics[width=1\textwidth]{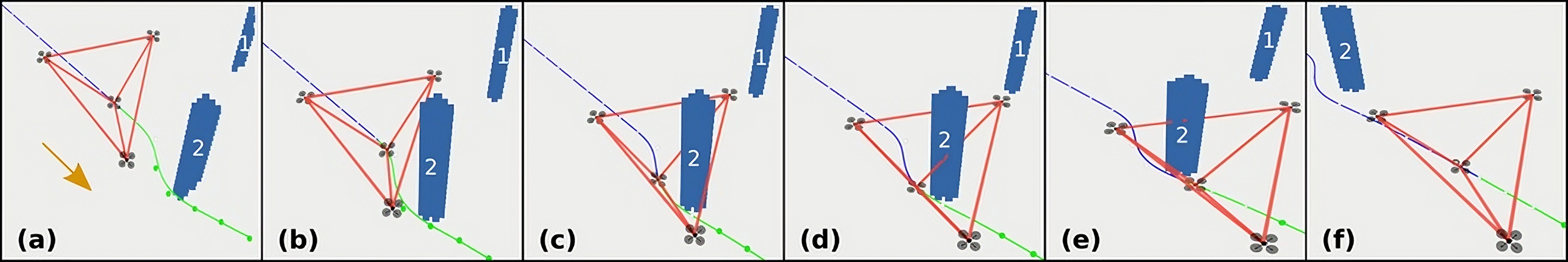}
    \caption{Snapshots of a 4-UAV swarm in a 3D formation (red lines), avoiding obstacles - presented with blue tiles and marked (1) and (2). The yellow arrow points to the direction of motion. The trajectory of the central UAV is re-planned (green line) based on the presence of the obstacles. The blue dotted line marks the path that has already been completed. (a) The swarm, still in formation, is approaching the obstacle. (b)-(e) The swarm formation brakes to prevent collision with the obstacle in its path. (f) The swarm returns to the initial formation after evading the obstacle.}
    \vspace{-2.5mm}
    \label{fig:system_overview.png}
\end{figure*}

For a UAV swarm system, the process begins with defining the area of interest and identifying critical Points of Interest (POIs), which are then prioritized. Suitable UAVs and sensors are selected to ensure an efficient swarm formation. A communication framework is established to facilitate seamless data exchange between UAV agents and the base station (see Fig.\,\ref{fig:system_design.png}).

\textit{Points (or area) of interest selection:}
Disaster management teams require critical information on environmental conditions and infrastructure damage during a fire event. Certain POIs are prioritized based on their significance in assessing the impact and guiding emergency response. Additionally, the dynamic nature of a post-disaster environment imposes time constraints on the availability of this information. 

To compensate for these factors, the proposed system accounts: (a) the coordinates of POIs, (b) a priority-based prize value that quantifies the significance of the acquired information, and (c) the opening and closing times of each POI, which define the time window when the surrounding area is deemed a safe flight zone, ensuring that all available information is collected while minimizing the risk to the UAV swarm. In cases where visiting all POIs is infeasible, the algorithm prioritizes those that provide the most critical information, ensuring an optimal balance between data acquisition and operational safety.

\textit{UAV Agents and Swarm Formation:} 
Emergency units in central European cities respond within 5–10 minutes and function in teams with distinct roles.
Following the same strategy, a well-configured UAV swarm must be deployed rapidly, with each agent having discreet roles and carrying the appropriate sensors, to enhance disaster assessment, reduce response times, and improve responder safety. Sensor selection depends on the disaster type, with essential tools including thermal scanners, chemical detectors, and radiation/nuclear sensors.
To showcase our framework, each UAV is equipped with a stereo camera for imaging and depth sensing, an IMU, and GPS for global position estimation. Furthermore, dedicated software that handles perception, trajectory planning, and control incorporating UAV dynamics for optimized path execution is integrated. GPS and IMU data are fused into the state estimation module for odometry extraction. All software modules are detailed in Section IV.

The duration of a disaster and the size of the affected area determine the necessary number of human agents. Similarly, the UAV swarm composition, including the number of agents and onboard specialized sensors (thermal camera, gas sensor, humidity sensor, LiDAR, etc.), is determined by the experts operating on-site. 
Each UAV's flight duration and velocity are constrained by its payload and specifications, stored in a capabilities database. 
The swarm configuration is selected as a subset of this database using an empirical rule that accounts for sensor heterogeneity, minimum flight duration, and minimum velocity, both of which are constrained by the weakest agent. 
Once the UAV count is set, the swarm is organized into a spatial formation—structured as a polygonal shape in 2D or 3D space depending on the area of interest—to ensure maximum coverage with available sensors. A designated UAV at the geometric center, chosen for its extended flight duration and high velocity with minimal payload, is a reference for trajectory generation and formation maintenance without carrying specialized sensors, similar to how first responders operate under an incident commander.

\textit{Communication:} The central UAV acts as a communication hub, maintaining strong signal links within the swarm to ensure reliable data transmission by minimizing the distance between agents. A UDP communication bridge is implemented on the central UAV, which is well-suited for decentralized swarms due to its low latency and minimal communication overhead. This setup enables all agents to broadcast and exchange odometry, local trajectory, and heartbeat signals, with each UAV uniquely identified by an assigned ID. The ground station only transmits waypoints for global trajectory planning and an execution signal, with the central UAV serving as the reference. 

\section{AUTONOMOUS NAVIGATION FRAMEWORK}
\subsection{Local map and obstacle detection}
This work extends our previous research in \cite{Geladaris2023RealTL} by developing an incrementally built Euclidean Signed Distance Field (ESDF) map for each UAV in the swarm. 
Each UAV computes its distance and gradient relative to nearby obstacles by utilizing depth sensor data and GPS-based odometry (see Section II). Instead of generating a global environment map, as in \cite{fiesta}, this approach maintains a local submap focused on the UAV’s immediate surroundings, significantly reducing computational overhead by excluding data not required for the re-planning process. 
The swarm’s collective perception enhances local mapping by covering a broader area, while the compact map size enables fast and accurate environmental representation, even in dynamic conditions.

\subsection{Swarm formation maintenance}
    The formation of UAVs is modeled by an undirected graph, where each vertex $i$ represents a UAV with position vector $\mathbf{p}_{i} = [ x_i, y_i, z_i]$. Each edge $e_{ij}$ connects UAV Agent $i$ and UAV Agent $j$ using the Euclidean distance between agents.

    To achieve the corresponding swarm formation, a formation similarity metric has been adapted \cite{quan2022distributedswarmtrajectoryoptimization} as
    \begin{equation}\label{eq:10}
                f=\parallel\mathbf{\hat{L}}-\mathbf{\hat{L}}_{des} \parallel_{F}^{2}=tr(\mathbf{\hat{L}}-\mathbf{\hat{L}}_{des})^{T} (\mathbf{\hat{L}}-\mathbf{\hat{L}}_{des}) 
    \end{equation}

    where $tr(\cdot)$ denotes the trace of a matrix, $\hat{\mathbf{L}}$ is the symmetric normalized Laplacian of the current swarm formation, $\hat{\mathbf{L}}_{des}$ the counterpart of the desired formation and $\| \cdot \|_{F}$ denotes the Frobenius norm.
    This metric is invariant to geometric transformations since the corresponding graph is weighted by the (absolute) Euclidean distance between UAVs. Scaling invariance is achieved with the use of normalized graph Laplacian.
    
\subsection{Local Planning Optimization}
Using distance and gradient information from the ESDF map, we employ an optimization-based local trajectory replanning algorithm similar to \cite{roboticsGeladaris}. Our approach integrates the essential terms into the cost function \(L\) of Eq.\,\ref{eq:1} to ensure a feasible, collision-free path while minimizing unnecessary maneuvers. It also includes agent-specific terms for formation maintenance and intra-UAV collision avoidance.
The algorithm generates a smooth B-Spline trajectory across sequential 3D points, leveraging the local control properties of B-Splines for real-time replanning \cite{overview}.

\begin{equation} \label{eq:1}
L_{i} = L_{c_i} + L_{{ep}_i} + L_{s_i} + L_{{fs}_i} + L_{{fa}_i}
\end{equation}

The overall cost function is a sum of individual agent terms, comprising: a collision term (Eq.\,\ref{eq:2}) to penalize paths that intersect with obstacles, an endpoint error term (Eq.\,\ref{eq:3}) to maintain replanned and initial trajectory consistency, a soft limit (Eq.\,\ref{eq:4}) that constrains velocity, acceleration, and jerk to allow the use of an unconstrained optimization algorithm, a swarm formation similarity term (Eq.\,\ref{eq:5}) to preserve formation integrity, and a reciprocal avoidance term (Eq.\,\ref{eq:6}) to prevent UAV-to-UAV collisions.

\begin{equation}\label{eq:2}
    L_{c} = \lambda_c \int_{t_i}^{t_f} (diff(t))^2 \,dt
\end{equation}

Where:
\begin{equation*}
    diff(t) = \begin{cases}
       0 &\quad\text{if} \text{      } d(t)-\tau > 0\\
       d(t) - \tau &\quad\text{if} \text{      } d(t)-\tau \le0\\
    \end{cases}
\end{equation*}
\(\lambda_c\) is a weight parameter, \(d(t)\) is the distance to the nearest obstacle calculated using trilinear interpolation from the incrementally built ESDF map, and \(\tau\) is a user-defined threshold.

\begin{equation}\label{eq:3}
    L_{ep} = \lambda_p||p_{glob}(t_f)-p(t_f)||^2 + \lambda_v||v_{glob}(t_f)-v(t_f)||^2
\end{equation}
\(\lambda_i\) are weight parameters for the position and velocity terms, \(p_{glob}(t_f)\) and \(v_{glob}(t_f)\) are the desired position and velocity respectively at the end of the time segment and \(p(t_f)\) and \(v(t_f)\) are the position and velocity respectively at the end of the segment of the B-Spline.

\begin{equation}\label{eq:4}
    L_{s} = \sum_{n=2}^{n=4} \int_{t_i}^{t_f} b(t,i) \,dt
\end{equation}

Where:
\begin{equation*}
    b(t,i) = \begin{cases}
       0 &\quad\text{if} \text{   } p^{(i)}(t) \le p_{max}^{(i)}\\
       exp((p^{(i)}(t))^2 - (p_{max}^{(i)})^2) &\quad\text{if} \text{   } p^{(i)}(t) > p_{max}^{(i)}\\
    \end{cases}
\end{equation*}
\(p^{(i)}(t)\) is the \textit{i-th} derivative of the position at time \textit{t} and \(p_{max}^{(i)}\) is the limit value.

\begin{equation}\label{eq:5}
    L_{fs}= \lambda_{fs} \sum_{j=0}^{K} \epsilon_f(t + jT)
\end{equation}

Where:
\begin{equation*}
\epsilon_{f} (t) =f ( p ( t ), \bigcup_{\Phi} p_{\phi} (t) ) 
\end{equation*}
is the similarity error between the current formation and the desired formation, and \(f(\cdot)\) is the function described in Eq.\,\ref{eq:10}. \(\Phi\) represents the collection of other agents. \( \lambda_{fs} \) is weight parameter for formation similarity term.

\begin{equation}\label{eq:6}
    L_{fa} = \lambda_{fa} \sum_{j=0}^{K} \sum_{\Phi} \max \{ \epsilon_{r_\phi}^{3}(p(t+jT), 0 \} 
\end{equation}

Where:
\begin{equation*}
\epsilon_{r_{\phi}} (t)=D_{r}^{2}-\parallel p ( t )-p_{\phi} (t) \parallel^{2}
\end{equation*}
\(D_{r}\) is the clearance distance from trajectories of other agents.

The cost terms described in Eq.\,\ref{eq:5} and Eq.\,\ref{eq:6} involve the expected trajectories of all the agents in the swarm. To implement them, both the number of B-spline control points \(K\) and spline segment time \(T\) must be considered.

\subsection{Traversal algorithms}

\begin{figure*}[t]
    \centering
    \includegraphics[width=1\textwidth]{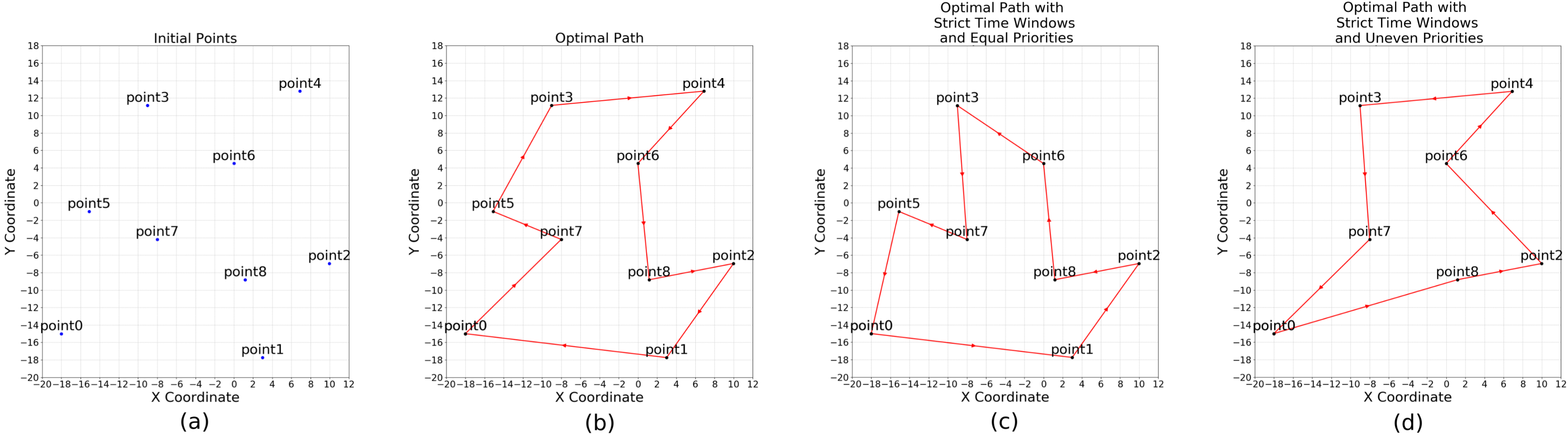}
    \caption{POIs sorted with the TSP algorithm, based on distance, prize collection, and time windows (TWs). (a) The initial list of POIs. (b) The optimal path with relaxed TWs and equal prize values (prize=10) for all POIs. The swarm can visit all POIs inside the given TW and collect all prizes. (c) The optimal path with strict TWs and equal prize values (prize=10) for all POIs. The swarm visits the maximum number of POIs in the available TWs to collect the maximum prize possible. (d) The optimal path with strict TWs and uneven prize values (prize=20 for Point 4, prize=5 for Point 5 and prize=10 for the rest). The swarm visits the POIs, attempting to collect the maximum prize possible in the given TWs.}
    \label{fig:POIs.png}
\end{figure*}

\begin{figure*}[t]
    \centering
    \includegraphics[width=1\textwidth]{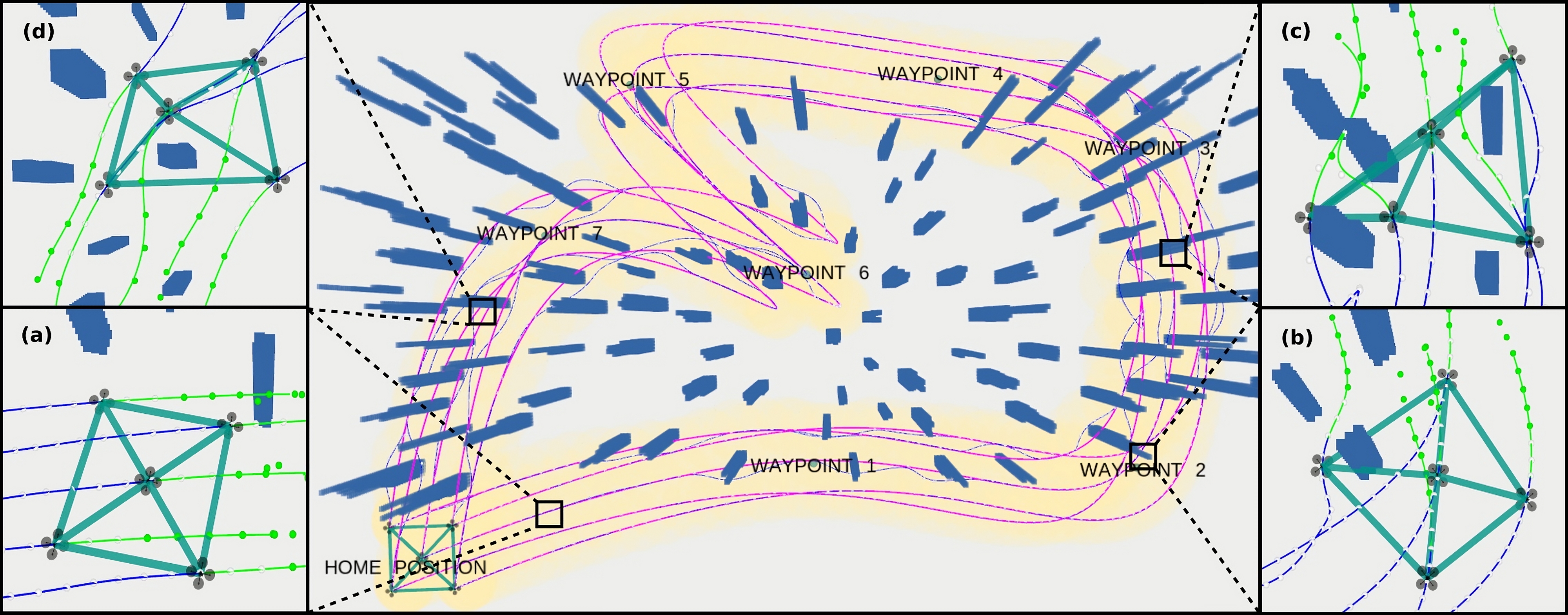}
    \caption{A squared-shaped formation (dark green lines) consisting of 5 UAVs traverses the global path (pink line) from the prioritized POIs, starting from a home position. Snapshots display swarm behavior in different timestamps. Trajectory replanning is able to provide obstacle-free trajectories (green line) while maintaining the initial formation. The total area coverage is highlighted in yellow, the obstacles are presented as blue tiles, and the path already covered by each agent is shown by a blue line. The green and white dots on the paths are the control points of the B-splines of the future and past paths accordingly. (a) The drone is in an area with sparse obstacles and maintains the formation. (b)-(d) The swarm moves along an area cluttered with obstacles and the formation changes according to the generated paths.}
    \label{fig:simulation_results}
\end{figure*}

To minimize transportation costs, the POIs are not visited randomly; instead, the final trajectory is computed as a solution to the TSP. 
To address this problem a variant of the problem known as the Prize-Collecting Traveling Salesman Problem (PC-TSP) is employed. In this approach, each POI is assigned a prize value and the swarm must visit a subset of points to maximize profit under constraints. This method allows for skipping less critical nodes, balancing the trade-off between travel costs and the rewards associated with visited points. The PC-TSP used in this work is specifically adapted to incorporate constraints related to infrastructure importance, the necessity of the required information, available flight time, and the capabilities of the onboard sensors. Furthermore, Time Windows (TW) are introduced, requiring each point to be visited within a specific time interval to ensure the feasibility of visits. 
The problem is formulated as a Mixed-Integer Linear Program (MILP), allowing for additional constraints while guaranteeing optimal solutions for up to 50 POIs. For larger POI sets, optimal solutions cannot be guaranteed due to computational time constraints on route optimization. However, these large sets can be hierarchically decomposed into smaller clusters to ensure tractable and efficient planning.

Two input tables are generated for the algorithm\footnote{\url{https://github.com/pigna90/PCTSPTW/tree/master}}. The first table specifies the opening and closing times of the time windows for each POI, along with their corresponding prize. The second is a symmetric table that records the cost of transport between each pair of POIs in time units, including the service duration. The open-source MILP solver CBC\footnote{\url{https://github.com/coin-or/Cbc}} computes the optimal visiting sequence, reporting total prizes collected and transport costs. 
The generated sequence is then converted to a global polynomial trajectory using the method presented in \cite{Richter2016}. 

\section{EXPERIMENTAL RESULTS}

To test the performance of our approach, a swarm of five UAVs in a square formation is commanded to navigate through a simulation environment with $110$ cylinder obstacles in an area of $60\,m\times25\,m$. Simulations were run on a 2.6\,GHz Intel i7 CPU with 16\,GB RAM. For trajectory execution, a geometric tracking controller with a multi-rotor dynamics model is used \cite{Furrer2016} and the ground truth odometry of the UAV agents is generated by the same quadrotor dynamics model.

As depicted in Fig.\,\ref{fig:simulation_results}, for a total of 10 experiments, the swarm traverses all POIs, successfully avoids all obstacles, and preserves the desired formation along the flight. The swarm formation error is computed as the difference of the Euclidean norm of the distance between the i-th and the central drone and their initial distance during the swarm's trajectory (Eq.\,\ref{eq:form_error}).

\begin{equation}\label{eq:form_error}
    \bar{\bm e}_{form} = \frac{\sum_{n=0}^{N}(||p^{(i)} - p^{(c)}||-d_0^{(i)})}{N}
\end{equation}

Table\,\ref{tab:formation_deviation} presents the mean and maximum deviations of each UAV from the initial swarm formation, using the central agent (UAV\,1) as a reference. These deviations are measured in scenarios where a five-agent swarm flies in a square formation with an edge length of 4 meters. Given that the radius of the obstacles is 0.5\,m\,-\,0.8\,m and the radius of the UAVs is 0.3\,m, the swarm maintains formation integrity while navigating cluttered environments, despite the impact of obstacle avoidance.

The trajectory error is computed as the Euclidean norm of the points of the initial and the re-planned trajectory (Eq.\,\ref{eq:traj_error}).

\begin{table}[b]
    \centering
    \begin{tabular}{|c|c|c|c|c|c|}
        \hline
        \textbf{Drone} & \textbf{UAV\,1} & \textbf{UAV\,2} & \textbf{UAV\,3} & \textbf{UAV\,4} & \textbf{UAV\,5} \\
        \hline
        \textbf{Mean (m)} & 0.000 & 0.359 & 0.423 & 0.327 & 0.363 \\
        \hline
        \textbf{Mean (\%)} & 0 & 12.7 & 14.9 & 11.6 & 12.8 \\
        \hline
        \textbf{Max (m)} & 0.000 & 2.558 & 3.294 & 1.590 & 2.053 \\
        \hline
    \end{tabular}
    \caption{Formation Deviation Metrics for Each UAV}
    \label{tab:formation_deviation}
\end{table}

\begin{equation}\label{eq:traj_error}
    \bar{\bm e}_{traj} = \frac{\sum_{n=0}^{N}||p_{g}^{(i)} - p_{r}^{(i)}||}{N}
\end{equation}

Table\,\ref{tab:trajectory_deviation} summarizes the deviations from the initial path for each UAV at experiments where the mean path length was 159.2\,m.
The swarm follows collision-free trajectories in cluttered environments while avoiding unnecessary maneuvers as the mean deviation of the UAVs from the initial path is always under 1.5\,m, when: $radius_{uav}+max(radius_{obs}) = 1.1\,m$.

\begin{table}[t]
    \centering
    \begin{tabular}{|c|c|c|c|c|c|}
        \hline
        \textbf{Drone} & \textbf{UAV\,1} & \textbf{UAV\,2} & \textbf{UAV\,3} & \textbf{UAV\,4} & \textbf{UAV\,5} \\
        \hline
        \textbf{Mean (m)} & 0.121 & 0.188 & 0.157 & 0.153 & 0.134 \\
        \hline
        \textbf{Max (m)} & 1.427 & 1.364 & 1.333 & 1.127 & 1.074 \\
        \hline
    \end{tabular}
    \caption{Trajectory Deviation Metrics for Each UAV}
    \label{tab:trajectory_deviation}
\end{table}

The efficiency of area coverage is quantified as the ratio of the actual area covered by the UAV swarm to the theoretical coverage achieved if waypoints were connected by direct straight-line paths. Given the TSP-generated paths, the swarm achieves full (100\,\%) spatial coverage of the expected area without gaps.

\section{CONCLUSIONS}

We have presented a complete framework for effectively leveraging a swarm of UAV agents capable of deploying heterogeneous sensor suites to cover an area of interest in disaster-affected regions. 
We showed that our system can avoid obstacles in cluttered environments while avoiding unnecessary maneuvers and maintaining swarm formation integrity, using distance and gradient information from an ESDF map. 
To achieve maximum area coverage for data collection, we formulated a TSP, with key infrastructure components identified as points of interest and prioritized. To ensure mission completion, we added time constraints in the TSP for efficiency and robustness. In the future, we plan to deploy our system in a real-world robotic swarm of aerial vehicles. This will allow us to validate the efficiency of our software across multiple UAVs, and enable robust task execution in even more complex and dynamic scenarios. Finally, more experiments will be conducted to benchmark our method against similar approaches.




\section*{ACKNOWLEDGMENTS}

This project has received funding from the European Union’s Horizon Europe programme under Grant Agreement N°101074008. 
The authors would like to thank Benjamin Schuster from the Johanniter Research \& Innovation Centre (Vienna, AT) for his valuable contribution to this work.
\renewcommand*{\bibfont}{\footnotesize}
\printbibliography


\end{document}